# Nanorobotics in Medicine: A Systematic Review of Advances, Challenges, and Future Prospects

Shishir Rajendran[1], Prathic Sundararajan[2], Ashi Awasthi[2], Suraj Rajendran[3]

*[1]Alliance Academy for Innovation, Cumming, GA 30040*
*[2]Wallace H. Coulter Department of Biomedical Engineering, Georgia Institute of Technology, Atlanta, GA 30332, USA*
*[3]Institute for Computational Biomedicine, Department of Physiology and Biophysics, Weill Cornell Medicine of Cornell University, New York, NY, USA*

**Abstract**
Nanorobotics offers an emerging frontier in biomedicine, holding the potential to revolutionize diagnostic and therapeutic applications through its unique capabilities in manipulating biological systems at the nanoscale. Following PRISMA guidelines, a comprehensive literature search was conducted using IEEE Xplore and PubMed databases, resulting in the identification and analysis of a total of 414 papers. The studies were filtered to include only those that addressed both nanorobotics and direct medical applications. Our analysis traces the technology's evolution, highlighting its growing prominence in medicine as evidenced by the increasing number of publications over time. Applications ranged from targeted drug delivery and single-cell manipulation to minimally invasive surgery and biosensing. Despite the promise, limitations such as biocompatibility, precise control, and ethical concerns were also identified. This review aims to offer a thorough overview of the state of nanorobotics in medicine, drawing attention to current challenges and opportunities, and providing directions for future research in this rapidly advancing field.

**Introduction**
Nanorobotics, a field merging nanotechnology with teleoperated and autonomous robotics, presents groundbreaking solutions that are unattainable with conventional robotics. A nanorobot, also known as a nanomachine, is a miniature mechanical or electromechanical device designed to perform specific tasks at the nanoscale level [1]. Contrary to nanorobotics, nanoparticles are tiny particles with unique properties, used for applications like drug delivery. Nanorobotics involves designing molecular-scale robots for tasks such as targeted medical procedures. The former is about passive materials, while the latter introduces active, controllable machines at the nanoscale. These miniature robots, due to their size, offer unique opportunities for operations at molecular and cellular levels.

The trend toward miniaturization in medical robotics has been gathering considerable momentum, and the potential impacts of this trend on the field of biomedicine are profound. Beyond the realm of macroscale medical robotics, the exploration of small-scale medical robotics, ranging from several millimeters down to a few nanometers in all dimensions, has intensified. These micro and nanoscale robots have been investigated for diverse biomedical and healthcare applications, including single-cell manipulation and biosensing, targeted drug delivery, minimally invasive surgery, medical diagnosis, tumor therapy, detoxification, and more [2].

By providing innovative ways to interact with biological systems at the cellular level, nanorobots promise to revolutionize various sectors of medicine, from diagnostics to treatment. The unique capabilities of nanorobots have opened up a new paradigm for problem-solving in biomedicine, enabling innovative approaches to challenges that were previously insurmountable. The potential to precisely manipulate biological materials at a cellular level has expanded the horizons of diagnostic and therapeutic procedures, bringing forth solutions that are more targeted, efficient, and minimally invasive.

This paper conducts a systematic review of the use of nanorobots in the medical field, with a specific focus on their applications and limitations in cancer treatment, dentistry, and cell therapy. We recognize that the field is continually evolving, and while our focus is on these three areas at present, the scope of this review could be adjusted to encompass emerging trends and breakthroughs. Through this exploration, we aim to provide a comprehensive overview of the current state of nanorobotics in medicine, trace the trajectory of this transformative technology, and highlight key challenges and potential solutions, providing direction for future research in this exciting and rapidly developing field.

**Methodology**
To obtain a comprehensive understanding of the applications and limitations of nanorobots in the medical field, we conducted a systematic review of the literature following the PRISMA (Preferred Reporting Items for Systematic Reviews and Meta-Analyses) guidelines. Our review was carried out by two independent reviewers, each thoroughly examining the available literature. The objective of this process was to identify research that provides information on how nanorobots are aiding advancements in the medical field, such as through nano cell manipulation robots or micro-laparoscopic surgery.

We did not set a date range for our literature search, thereby including the earliest relevant papers on the topic to the most recent ones, aiming to capture the full development arc of nanorobotics in medicine. The selection criteria necessitated the literature to be inclusive of both nanorobotics and medicine. Therefore, articles solely focusing on nanorobots without any direct medical applications, or articles strictly on the medical field without reference to nanorobots were excluded from our review. To achieve this, we employed a structured keyword search strategy, outlined in the next section, which guided the process of literature selection, ensuring a balanced representation of the two intersecting domains under review: nanorobotics and medicine.

The literature search was conducted on two comprehensive and well-regarded databases: IEEE Xplore and PubMed. These databases were chosen based on their accessibility and the high merit of the works they contain, particularly in the intersecting domains of technology and medicine. In analyzing the data, we noticed an increase in the number of papers published over time on this topic, indicating growing interest in the intersection of nanorobotics and medicine. This trend is visually represented in figure 2, underscoring the escalating attention given to this research area. Our search was broad, without any date range filters, in order to capture the evolution and growth of the field since its inception. The keyword search was constructed to pull relevant literature that overlapped the areas of nanorobotics and biomedicine. Our general keyword search strategy was as follows: ("swarm robotics" OR "swarm intelligence" OR "swarm behavior" OR "Multi-robot systems" OR "microbots" OR "nanorobots") AND ("biomedicine" OR "medical applications" OR "healthcare" OR "medicine" OR "surgery"). The same

keywords were used for the IEEE Xplore database. For the PubMed search, these terms were additionally constrained to the Title/Abstract fields of the papers. From our initial search, we retrieved 110 results from PubMed and 304 results from IEEE Xplore. These results underwent further screening and filtration based on the relevance and quality of the content, as detailed in the next section.

Following the initial extraction of papers, the literature was further refined through a second stage of filtration based on their relevance to both nanorobotics and medical applications. This involved a detailed review of each paper to ensure that they indeed intersected these two topics. Papers that focused exclusively on either nanorobots (without direct medical application) or medical topics (without the use of nanorobots) were excluded at this stage. During the filtration process, we also noticed a considerable number of papers that focused on 'nano intelligence' rather than 'nano robotics' or 'nano technology'. While related, the domain of 'nano intelligence' largely covers algorithmic development and computational models, which falls outside the mechanical or electromechanical focus of nanorobotics. Considering the scope and purpose of our review, we decided to exclude these papers to maintain a clear focus on nanorobotics in the context of medical applications. That being said, we have included some broader literature on the general topic of nanorobotics, which provided essential context and historical development of the field. This allows us to present a comprehensive picture of the journey of nanorobots from a conceptual stage to the sophisticated tools they represent today in the realm of medicine. A full pipeline of our methodology can be found in **Supplemental Figure 1**.

**Results**

The stringent filtration process resulted in a final selection of 52 papers that fit our criteria. An analysis of these papers revealed their focus on diverse subfields within the scope of nanorobotics in medicine. Specifically, 15 papers were dedicated to cancer-related research, 7 papers targeted cell, tissue, or organ treatment, 12 papers discussed surgical applications, 8 papers covered nanorobotic applications in drug delivery and 5 papers focused on applications in dentistry. Refer to **Figure 1**. The remaining papers consisted of general reviews on nanorobotics or tackled miscellaneous topics that could not be neatly categorized into any of the aforementioned areas.

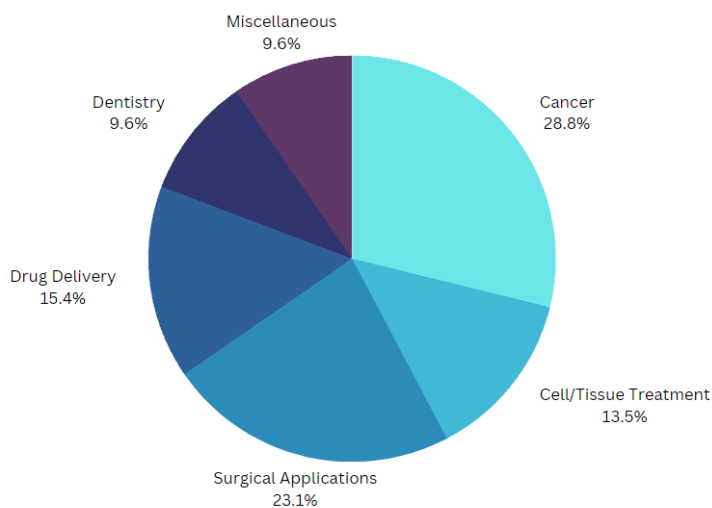

**Figure 1. Pie Chart of Applications of Nanotechnology in Medicine**

A temporal analysis of these selected papers indicated a notable trend: out of the 52 papers, 35 were published after the year 2017, and the majority of these appeared after 2021. This pattern not only signals a burgeoning interest in the field but also points to the rapid evolution of nanorobotics in medical applications in recent years. This trend is depicted graphically in **Figure 2** of the growing scholarly attention to this field over time.

Interestingly, the selected papers showcased a high degree of international authorship, pointing to the global interest and collaborative effort in exploring nanorobotic applications in medicine. We provide a graphical representation of this international engagement in **Figure 3** below, further underscoring the widespread academic pursuit of solutions and advancements in the field of nanorobotics.

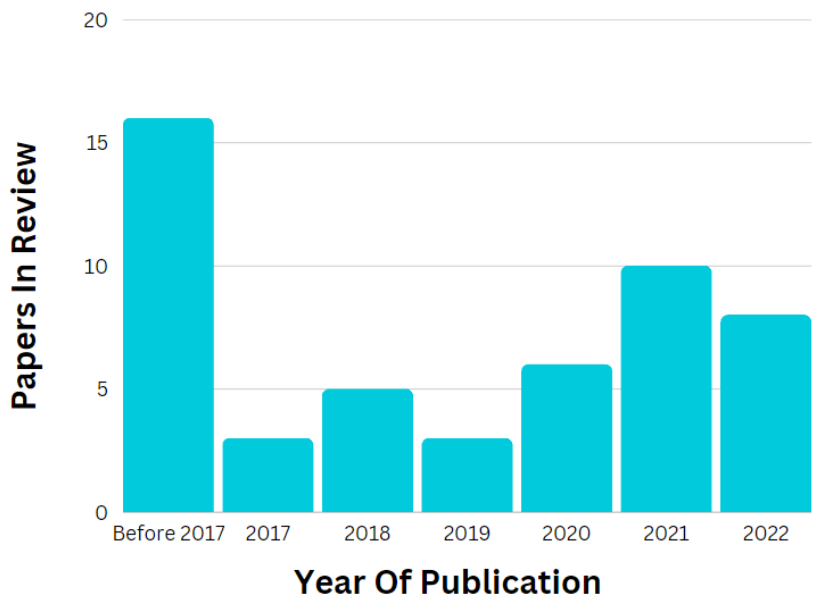

**Figure 2. Number of papers in the field by year.** y-axis shows the number of papers published and x-axis is the year(s) of publication

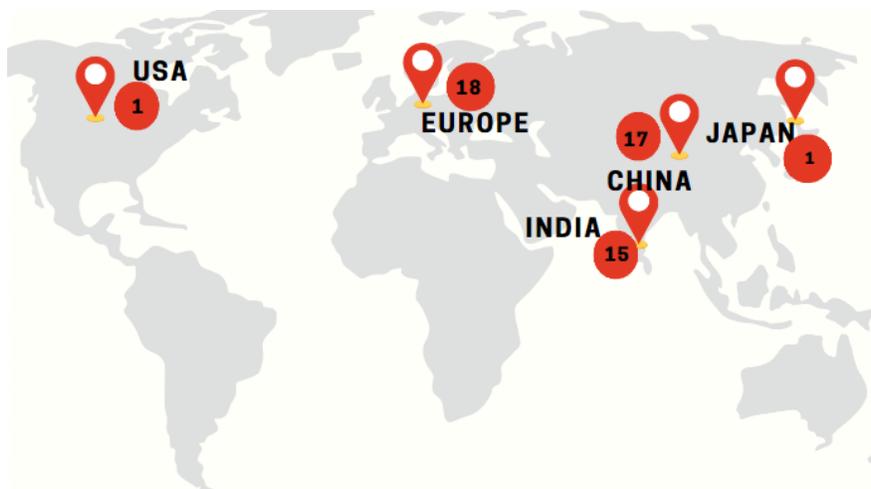

**Figure 3. Number of papers in biomedical nanorobotics by region.** Number inside each red circle represents the quantity of papers from each region.

## Discussion
*Cancer*

Cancer remains one of the most formidable health challenges worldwide. Its complexity and heterogeneity necessitate treatments that can target aberrant cells specifically while sparing healthy ones.

Over the years, traditional treatments like radiation and chemotherapy have seen widespread use; however, their systemic administration often leads to significant adverse effects due to a lack of selectivity. Herein, the field of nanorobotics offers an innovative paradigm for cancer therapeutics.

Nanorobots, characterized by their operation at the molecular level, are well-suited for cancer treatment applications. Their miniature size, comparable to biological macromolecules, allows them to navigate the intricate biological landscape with a degree of precision unattainable by traditional therapeutic modalities. This high level of precision helps in reducing the harm to non-target cells, a notable downside in the standard radiation and chemotherapy treatments. One of the critical utilities of nanorobots in oncology is the provision of high-resolution information to surgeons. Through their ability to map out cancerous cells in site, nanorobots can aid surgeons in planning and executing intricate surgical procedures, thereby enhancing surgical outcomes and patient prognosis. For instance, during laparoscopic cancer surgery, nanorobots could offer real-time mapping of the areas requiring dissection, thereby guiding the surgeon's actions to maximize tumor removal and minimize damage to healthy tissues. Moreover, beyond their navigational prowess, nanorobots have the potential to directly intervene in the tumor microenvironment. They can carry therapeutic payloads, specifically deliver these to cancer cells, and even execute programmable actions upon reaching the target site. This feature helps to significantly increase the treatment's specificity, subsequently reducing systemic toxicity [3], [4].

The temporal aspect of cancer treatment is a critical consideration, often linked to patient survival rates. Traditional treatment methods like chemotherapy, while effective, are often protracted, requiring multiple cycles spread over months. In contrast, the precise delivery and operation of nanorobots can expedite the treatment process, potentially resulting in quicker therapeutic responses. The versatility of nanorobotics in cancer therapy is further exemplified by their potential synergy with conventional treatment modalities. Nanorobots can also serve as reservoirs of therapeutic agents in the bloodstream [5]. By continuously releasing agents, such as Doxorubicin and Paclitaxel (used for chemotherapy), over an extended period, they can enhance the efficacy of chemotherapy and potentially other systemic treatments. This ability of nanorobots to act as delivery platforms can significantly extend the functional half-life of the therapeutic agents and maintain optimal drug concentration levels in the systemic circulation. By operating in tandem with traditional therapies such as chemotherapy or radiation, nanorobots can augment the overall therapeutic outcomes while potentially mitigating side effects through targeted delivery.

An integral feature of nanorobots that considerably enhances their functionality in cancer treatment is their sensor-based ability to detect and surgically excise tumors. Innovative frameworks such as Tumor Sensitization and Targeting (TST) have been proposed which aim to aid surgeons in detecting and operating on tumors located within difficult-to-reach tissues and body cavities, a task currently beyond the capabilities of existing surgical technologies [6], [7]. What makes these nanorobotics-based strategies unique is their reliance on swarm intelligence. A large number of nanorobots can work collectively, like a swarm, to achieve a common goal [8]. This approach leverages the power of cooperation, enabling nanorobots to cover large areas or perform complex tasks that would be impossible for a single nanorobot. Moreover, the design and fabrication of nanorobots have also been inspired by naturally occurring biological substances. These biomimetic nanorobots incorporate natural components to circumvent potential immune responses and to minimize side effects during treatment. By aligning with the body's innate biological systems, these nanorobots can function more efficiently, reducing the potential for adverse reactions while maintaining therapeutic effectiveness. These approaches encompass the emulation of red blood cell attributes within the bloodstream to mitigate immune reactions, and the strategic application of specialized coatings to evade immune system recognition.

An essential characteristic of nanorobots that significantly contributes to their effectiveness in cancer management is their ability to traverse freely within the body without causing disruptions or adverse effects [9]. Their small size, comparable to that of cellular organelles, is what permits them to operate at the cellular level, providing therapeutic benefits directly to the cells [10]. Their nanometric scale allows these machines to infiltrate cellular structures, perform tasks, and navigate the intricate pathways within the human body that are inaccessible to conventional medical tools. This capability has the potential to revolutionize treatment methods by providing highly localized treatments, which can lead to more efficient therapies with fewer side effects. Furthermore, due to their small size, nanorobots can be utilized in higher numbers to maximize their collective impact, bringing us back to the concept of swarm intelligence. By combining their individual efforts, nanorobots can collectively execute complex tasks more efficiently than a single entity, making them an invaluable tool in the medical field.

*Surgical Applications*
The use of nanorobots in surgical applications has demonstrated a substantial potential for enhancing the precision and efficiency of medical procedures. For instance, systems like SARAS (Smart Autonomous Robotic Assistant Surgeon), which were released in 2000 and 2016 respectively, have been developed to assist during surgeries by executing tasks that would otherwise burden the surgical team [11], [12], [13], [14]. As a result, the operation time is significantly reduced, allowing for more efficient use of resources and potentially leading to improved patient outcomes.

In a similar vein, RMIS (Robotic Minimally Invasive Surgery) systems such as the da Vinci Surgical System employ smaller nanorobots to execute delicate movements with precision. By supplementing human dexterity with robotic precision, these systems support surgeons in carrying out surgeries more efficiently and accurately [11], [14], [15].

These advanced systems have undergone extensive testing to evaluate their efficacy in assisting surgeons. [16]. The results show that nanorobot-assisted surgical systems can indeed provide substantial support to surgeons. They allow for a higher level of control and precision, lower invasiveness, and reduced operation times, representing a promising trend for the future of surgical procedures.

Nanorobots possess the unique ability to integrate with materials such as colloidal gold and quantum dots. These materials exhibit distinct structural and chemical properties that are not accessible to larger-scale entities [17]. This unique feature enhances the versatility of nanorobots, enabling them to perform tasks that would be challenging or even impossible for larger robots, thus making surgical procedures more efficient and manageable. For instance, nanorobots equipped with such materials can be used to target and treat tumors and malignant cells with a high degree of precision. These materials, when used in conjunction with nanorobots, can facilitate the targeted delivery of therapeutic agents in high specificity and/or affinity-type environments [18]. This ability to accurately target pathological cells while sparing healthy ones not only improves the efficacy of the treatment but also minimizes potential side effects, thereby showcasing the vast potential of nanorobots in advancing surgical applications.

Techniques such as the PANDA ring resonator utilize these distinct materials to transport nanorobots and enable their deployment for surgical treatment. Nanorobots employ a coating of colloidal gold, which facilitates their precise navigation and operation in complex biological environments [19].

In addition to material advancements, novel propulsion methods for nanorobots are being explored. Some of these techniques include the use of light-driven nanomotors. The strength of the motors can be adjusted by varying the intensity of light [20], [21]. This allows for the fine-tuning of nanorobot movement and control, critical for navigating intricate biological structures.

These innovations in nanorobotics open the door to previously unimaginable applications in biomedicine. It becomes conceivable that complex surgical procedures could be conducted at single-cell precision without the need for invasive surgical incisions. Nanorobots promise a future where medical interventions are not only less invasive but also more precise and personalized, reinforcing the value of nanorobotic research and development.

*Drug Delivery*
Nanorobots can revolutionize drug delivery methods, improving the speed, efficiency, and specificity of treating diseases and infections. By utilizing built-in sensors, nanorobots can precisely locate diseased or infected regions within the body, where the administration of therapeutic drugs is necessary. These nanorobots can autonomously or via remote controls administer the appropriate drugs at the target site, effectively bypassing the need for invasive surgical procedures [22], [23], [24].

The advent of nanorobotics in drug delivery can lead to significantly enhanced therapeutic outcomes, while minimizing potential side effects commonly associated with systemic drug delivery. By delivering drugs directly to the pathologic site, nanorobots can ensure that the therapeutic agents exert their maximal effect at the desired location while minimizing systemic exposure, thus reducing the likelihood of adverse effects. This concept, often referred to as targeted drug delivery, is one of the most promising benefits of integrating nanorobotics into modern medicine.

Janus micro/nanorobots, named for their dualistic nature, represent a promising frontier in drug delivery applications. A Janus micro/nanorobot is a mobile micro/nanomachine with a dual-structure that can efficiently transform various energy sources, including both local and external power, into mechanical force, encompassing motors, swimmers, and actuators, among others [25]. The advantageous structure of Janus robots enables them to leverage fuel-effective materials, which, coupled with their exceptional maneuverability, allows them to navigate within the human body with remarkable precision. Their dualistic nature, often constituted by one side being passive and the other active, facilitates differential responses to the environment or stimuli, providing them a controlled and directed motion capability. This precise control over their movement, when paired with appropriate drug-carrying materials, opens up new possibilities for drug delivery. They can transport therapeutics directly to the required site, thereby minimizing systemic side effects and maximizing drug efficacy. The efficient energy translation capabilities and precision navigation of Janus robots substantially enhance the efficacy of drug delivery. These nanoscale robots offer a revolutionary approach to delivering therapeutics, with potential applications extending beyond medicine into broader fields of biotechnology and nanotechnology.

Nanorobots present an all-in-one solution for drug delivery, demonstrating capabilities in sensing, initiating, and administering treatment in targeted areas. These miniature machines, due to their intricate design and multifaceted functionality, can serve various roles in the biomedical field [15], [26]. One key advantage of nanorobots is their ability to navigate through biological fluids, a task previously unachievable by larger-scale medical devices. Recent research has proposed fluid-traveling nanorobots, which use flagellar motion patterns—akin to the propulsion method used by certain bacteria—to travel within the human body. These nanorobots, essentially microscale swimmers, can traverse the viscous environment of bodily fluids, delivering therapeutics directly to the targeted site [27], [28].

By mimicking the biological propulsion mechanisms found in nature, these nanorobots can overcome the physical challenges presented by the human body's environment. Therefore, the ability of nanorobots to perform an array of tasks within a single unit—coupled with their unique propulsion mechanisms—heralds a new era in drug delivery and, more broadly, in the field of nanomedicine.

*Cellular Nanorobotics*
The compact size of nanorobots enables them to penetrate cells, opening up new possibilities for cellular treatments. These microscopic machines can be pre-programmed with specific functions, which reduces the likelihood of errors and broadens the scope of possible treatments [29].

One noteworthy type of nanorobot is the DNA nanorobot, which has been gaining significant attention in recent years. These nanorobots are engineered with a plethora of unique attributes, such as tissue penetration, site-targeting, stimuli responsiveness, and cargo-loading capabilities. This makes them highly suited for precision medicine applications, as they can deliver targeted interventions at a cellular or even molecular level. Nanorobots can be designed with sophisticated logic gates, enabling them to perform a series of actions in response to various stimuli [29], [30], [31]. This level of functional complexity enhances their versatility in biomedical applications and promises to pave the way for more personalized, efficient, and accurate treatments.

One of the significant advantages of nanorobots is their ability to be engineered using various functional nanomaterials, which can give them diverse functionality. The integration of these nanomaterials can modulate a nanorobot's performance, allowing it to adapt to a range of biomedical tasks. [32], [33], [34]. The flexibility of nanorobots' design and functionality is highlighted by the development of 'Respirocytes,' artificial red blood cells capable of carrying oxygen and carbon dioxide. These respirocytes can serve as temporary substitutes for natural blood cells during emergencies, thereby revolutionizing the treatment of heart diseases and the field of hematology in general [35], [36], [37].

Further, nanorobots have been demonstrated to be capable of promoting desirable cell behavior and growth. This means they can potentially regulate cell function, thanks to their versatility [38], [39]. These developments underscore the potential of nanorobots to serve as powerful tools for precision medicine, capable of operating at a cellular or even molecular level to diagnose, monitor, and treat diseases [40].

*Miscellaneous*

The collective operation of nanorobots, or swarm robotics, presents a transformative approach to a variety of other medical applications. Nanorobots functioning in a swarm can be coordinated to execute complex tasks collectively and cooperatively, which can significantly enhance their efficacy. Swarms of nanorobots can be maneuvered collectively via the same fuel source, such as light or magnetic fields [41], [42].

This ability for swarm control means that all the nanorobots can be simultaneously subjected to the same stimuli, streamlining the execution of medical procedures. Moreover, the application of swarm intelligence models can potentially simplify the control over these nanorobots swarms, making them more manageable [43]. The application of swarm intelligence extends to medical imaging, where it has been deployed to identify metastasis, micro-calcifications, and for brain image segmentation [44], [45], [46]. These findings underscore the potential for nanorobot swarms to revolutionize areas of medicine ranging from targeted therapeutics to sophisticated imaging techniques.

Nanorobots also offer promising advances in the field of medical instrumentation, owing to their versatility and adaptability. The construction of nanorobots can be adjusted through the manipulation of their constituent materials, thereby enabling them to serve diverse functionalities, which is an invaluable feature for medical instrumentation [47].

Neurology can also benefit from nanorobots, particularly in the precise delivery of neuroprotective drugs and the monitoring of neurological disorders. For instance, nanorobots could be engineered to transport drugs across the blood-brain barrier, a challenge that conventional methods often struggle with. This could facilitate the treatment of conditions like Parkinson's disease by delivering therapeutic agents directly to affected brain regions, thereby enhancing drug efficacy and minimizing side effects. Furthermore, nanorobots offer a promising avenue for monitoring and managing neurological disorders. By integrating sensors or imaging components into their design, nanorobots could provide real-time data on brain activity or the presence of specific biomarkers associated with disorders such as epilepsy. This would enable more accurate diagnoses and personalized treatment strategies [48].

The utility of nanorobots extends to orthopedics, ophthalmology, and infectious diseases as well. In orthopedics, nanorobots could be used for the regeneration of bone tissue or the precise delivery of drugs to a particular joint. In ophthalmology, nanorobots might assist in treating eye conditions at a cellular level, providing targeted treatments that could improve upon current methods. In the realm of infectious diseases, nanorobots could potentially be programmed to seek out and destroy specific pathogens, thereby providing a highly specific treatment strategy.

The application of nanorobotics in orthodontics is a noteworthy advancement in the dental field. It has revolutionized the way dental care is delivered by automating procedures that were previously manual, thereby reducing human error and increasing efficiency. Nanorobotics has a variety of applications in dental care including cleaning, whitening, and surgical procedures. One development in orthodontics is the use of Temporary Anchorage Devices (TADs), which includes devices such as miniscrews, miniplates, and implants. These devices are embedded into the bone structure and are designed to enhance orthodontic anchorage. They can either provide independent anchorage or bolster the anchoring teeth. The key advantage of using nanorobotics in this context is their ability to be programmed to perform precise tasks. This can lead to more efficient and effective dental procedures, particularly when dealing with complex orthodontic cases such as surgery or implants. Furthermore, nanorobotics allows for more localized treatment, reducing the chances of potential side effects and discomfort for patients [49], [50], [51]. Incorporating nanotechnology and nanorobots in dentistry has the potential to dramatically enhance the standard of care provided. Nanotechnology can be crucial in managing bacterial biofilms and aiding remineralization of teeth post-decay, thus promoting oral health and preventing further dental issues. A key application of nanotechnology in orthodontics is in reducing frictional forces in orthodontic systems. A proposed solution for this pain point is to coat orthodontic archwires with a film that incorporates nanoparticles with nanorobotic machines, which would make the system function more smoothly. This has the potential to make orthodontic treatments more comfortable and efficient for patients. Moreover, the versatility of nanorobots makes them suitable for use in virtually every aspect of dental care and treatment. They can be utilized for routine procedures such as cleaning, as well as for more specific applications such as cosmetic dentistry, teeth whitening, addressing hypersensitivity, and orthodontics. The use of nanotechnology in dentistry holds great promise. The potential to streamline processes, enhance patient comfort, and improve treatment outcomes are some of the many benefits we could see from the continued integration of nanorobots in this field. However, as with all emerging technologies, careful research and consideration are necessary to ensure the safety and efficacy of these innovations.

**Limitations**

The introduction of nanorobots into medical treatment confronts a multitude of intricate challenges and limitations, each requiring careful examination. A primary concern centers on the biocompatibility of nanorobots, especially when inorganic materials are integrated into their construction. Inadequate biocompatibility can trigger unwarranted immune responses, such as the recruitment of white blood cells, potentially compromising nanorobot functionality and posing health risks to the patient. Therefore, meticulous examination, testing, and optimization of nanorobot materials are essential to ensure their seamless integration with the human body. For instance, nanorobots employing biocompatible polymers or surface modifications, such as PEGylation, can enhance their biocompatibility [52].

High costs associated with nanorobot development constitute a multifaceted challenge. The precision and intricacy of manufacturing demand substantial resources, including state-of-the-art fabrication facilities and specialized expertise. Additionally, the programming of nanorobots to execute intricate tasks and their deployment in clinical settings contribute significantly to their expense. The cost-benefit analysis of nanorobot utilization must address these financial constraints, especially in healthcare systems with limited resources.

Effective integration of nanorobots with complementary medical techniques, such as medical imaging systems, is paramount. This integration enables precise navigation and real-time monitoring of nanorobots within the body. The complexity arises from the need for seamless coordination between nanorobot functions and imaging modalities. Additionally, the integration necessitates substantial resources, both in terms of equipment and expertise, to ensure seamless cooperation between these technologies [47], [52].

Administering nanorobots intravenously poses intricate challenges. As these diminutive agents traverse the bloodstream, they confront various barriers. These encompass the possibility of provoking immune responses, the potential for adverse side effects arising from interactions with diverse cell types, and heightened concerns about causing obstructions or damage to blood vessels during intricate navigation [5].

The financial implications of intravenous administration are notable, as precise manufacturing, the orchestration of sophisticated control mechanisms, and ongoing maintenance are necessary. Thus, scalability in resource-limited healthcare contexts is hampered by financial constraints. The endurance of nanorobots under physiological conditions is a substantial concern. Questions loom over their ability to function optimally over extended durations. This uncertainty could necessitate repeated dosing regimens, further increasing the financial burdens associated with nanorobot deployment. Strategies like incorporating resilient materials and developing self-repair mechanisms are under investigation to address these concerns.

The quest for safe and efficient propulsion methods within the human body is a complex challenge. Propulsion mechanisms must meet stringent criteria, including non-toxicity, non-immunogenicity, and nanoscale efficiency, to ensure both safety and efficacy. Promising methods include using biological motors, such as flagella, to propel nanorobots, but extensive research is ongoing to optimize these systems for clinical use. Additionally, regulatory frameworks for nanorobots in medical contexts are still in their infancy. A comprehensive and stringent regulatory framework is indispensable to ensure the safety and efficacy of nanorobotic devices. This framework necessitates exhaustive pre-clinical testing, validation, and risk assessment, adding a layer of complexity and cost to the deployment process. International standards are also under development to harmonize regulations across borders. These multifaceted and highly nuanced challenges underscore the necessity for interdisciplinary collaboration spanning engineering, biology, ethics, and regulatory oversight to navigate the intricacies and address the limitations associated with nanorobots, ultimately paving the way for their effective and responsible integration into medical practice.

**Future Directions**
Nanorobotics is a rapidly advancing field with the potential to revolutionize medicine in a myriad of ways. As a promising tool, nanorobots hold the potential to enhance drug delivery, enable precise surgical interventions, promote desired cellular behavior, and even contribute to routine dental care. While it's true that there are still obstacles to overcome, including bio incompatibility, high costs, and the challenges of independent functionality, the advancements in nanotechnology and robotics are paving the way for potential solutions. As our understanding of nanoscale materials, propulsion mechanisms, and cellular interactions deepens, it is likely that these challenges will become less formidable. Moreover, as resources become increasingly available, and as the designs for various nanorobots with different functionalities continue to evolve, it is expected that these tiny medical marvels will find wider applications in the medical field. Growing interest in this field will likely spur further research and development, leading to more innovative designs and applications. Future work should focus on making nanorobots more biocompatible, cost-effective, and autonomous, as well as on developing safe and efficient propulsion methods. Furthermore, the integration of nanorobots with medical imaging systems should be explored in greater depth. Collaboration across disciplines, such as materials science, biology, robotics, and medicine, will be essential to realize the full potential of nanorobots. With more research, regulatory advancement, and clinical trials, we can anticipate a future where nanorobots are a standard part of medical treatments, improving outcomes, and quality of life for patients.

**Declaration of Interests**
The authors have no competing interests.

**Supplemental Information**
*Supplemental Figure 1*. PRISM Diagram illustrating article selection for systematic review

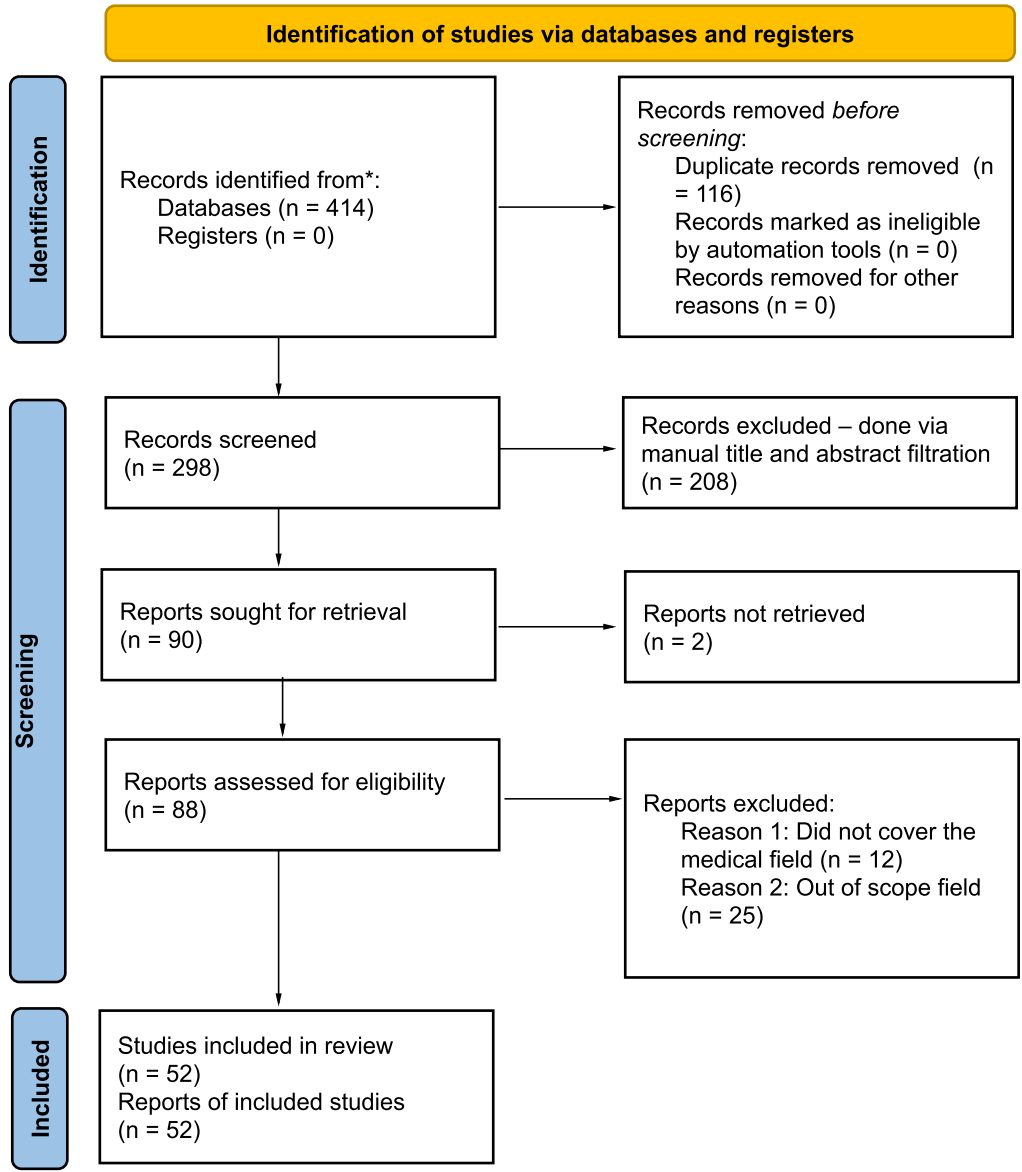

**References**

[1] Patel GM, Patel GC, Patel RB, Patel JK, Patel M. Nanorobot: a versatile tool in nanomedicine. J Drug Target. 2006 Feb;14(2):63-7. doi: 10.1080/10611860600612862. PMID: 16608733.

[2] Li M, Xi N, Wang Y, Liu L. Progress in Nanorobotics for Advancing Biomedicine. IEEE Trans Biomed Eng. 2021 Jan;68(1):130-147. doi: 10.1109/TBME.2020.2990380. Epub 2020 Dec 21. PMID: 32340931.

[3] M. Venkatesan and B. Jolad, "Nanorobots in cancer treatment," INTERACT-2010, Chennai, India, 2010, pp. 258-264, doi: 10.1109/INTERACT.2010.5706154.

[4] A. Cavalcanti, B. Shirinzadeh, D. Murphy and J. A. Smith, "Nanorobots for Laparoscopic Cancer Surgery," 6th IEEE/ACIS International Conference on Computer and Information Science (ICIS 2007), Melbourne, VIC, Australia, 2007, pp. 738-743, doi: 10.1109/ICIS.2007.138.

[5] Aggarwal M, Kumar S. The Use of Nanorobotics in the Treatment Therapy of Cancer and Its Future Aspects: A Review. Cureus. 2022 Sep 20;14(9):e29366. doi: 10.7759/cureus.29366. PMID: 36304358; PMCID: PMC9584632.

[6] Shi S, Yan Y, Xiong J, Cheang UK, Yao X, Chen Y. Nanorobots-Assisted Natural Computation for Multifocal Tumor Sensitization and Targeting. IEEE Trans Nanobioscience. 2021 Apr;20(2):154-165. doi: 10.1109/TNB.2020.3042266. Epub 2021 Mar 31. PMID: 33270565.

[7] Deng G, Peng X, Sun Z, Zheng W, Yu J, Du L, Chen H, Gong P, Zhang P, Cai L, Tang BZ. Natural-Killer-Cell-Inspired Nanorobots with Aggregation-Induced Emission Characteristics for Near-Infrared-II Fluorescence-Guided Glioma Theranostics. ACS Nano. 2020 Sep 22;14(9):11452-11462. doi: 10.1021/acsnano.0c03824. Epub 2020 Aug 26. PMID: 32820907.

[8] Y. Chen, Y. Wang, L. Cao and Q. Jin, "An Effective Feature Selection Scheme for Healthcare Data Classification Using Binary Particle Swarm Optimization," 2018 9th International Conference on Information Technology in Medicine and Education (ITME), Hangzhou, China, 2018, pp. 703-707, doi: 10.1109/ITME.2018.00160.

[9] Sushmitha Nehru, Ranjita Misra, and Maharshi Bhaswant ACS Biomaterials Science & Engineering 2022 8 (2), 444-459 DOI: 10.1021/acsbiomaterials.1c01352

[10] Li J, Esteban-Fernández de Ávila B, Gao W, Zhang L, Wang J. Micro/Nanorobots for Biomedicine: Delivery, Surgery, Sensing, and Detoxification. Sci Robot. 2017 Mar 15;2(4):eaam6431. doi: 10.1126/scirobotics.aam6431. Epub 2017 Mar 1. PMID: 31552379; PMCID: PMC6759331.

[11] A. Leporini et al., "Technical and Functional Validation of a Teleoperated Multirobots Platform for Minimally Invasive Surgery," in IEEE Transactions on Medical Robotics and Bionics, vol. 2, no. 2, pp. 148-156, May 2020, doi: 10.1109/TMRB.2020.2990286.

[12] Mertz L. Tiny Conveyance: Micro- and Nanorobots Prepare to Advance Medicine. IEEE Pulse. 2018 Jan-Feb;9(1):19-23. doi: 10.1109/MPUL.2017.2772118. PMID: 29373853.

[13] M. Deniša, K. L. Schwaner, I. Iturrate and T. R. Savarimuthu, "Semi-Autonomous Cooperative Tasks in a Multi-Arm Robotic Surgical Domain," 2021 20th International Conference on Advanced Robotics (ICAR), Ljubljana, Slovenia, 2021, pp. 134-141, doi: 10.1109/ICAR53236.2021.9659445.

[14] R. Konietschke, T. Ortmaier, U. Hagn, G. Hirzinger and S. Frumento, "Kinematic Design Optimization of an Actuated Carrier for the DLR Multi-Arm Surgical System," 2006 IEEE/RSJ International Conference on Intelligent Robots and Systems, Beijing, China, 2006, pp. 4381-4387, doi: 10.1109/IROS.2006.282014.

[15] Menciassi A, Sinibaldi E, Pensabene V, Dario P. From miniature to nano robots for diagnostic and therapeutic applications. Annu Int Conf IEEE Eng Med Biol Soc. 2010;2010:1954-7. doi: 10.1109/IEMBS.2010.5627629. PMID: 21097006.

[16] D. Ceraso and G. Spezzano, "Controlling swarms of medical nanorobots using CPPSO on a GPU," 2016 International Conference on High Performance Computing & Simulation (HPCS), Innsbruck, Austria, 2016, pp. 58-65, doi: 10.1109/HPCSim.2016.7568316.

[17] Singhal S, Nie S, Wang MD. Nanotechnology applications in surgical oncology. Annu Rev Med. 2010;61:359-73. doi: 10.1146/annurev.med.60.052907.094936. PMID: 20059343; PMCID: PMC2913871.

[18] Kishore C, Bhadra P. Targeting Brain Cancer Cells by Nanorobot, a Promising Nanovehicle: New Challenges and Future Perspectives. CNS Neurol Disord Drug Targets. 2021;20(6):531-539. doi: 10.2174/1871527320666210526154801. PMID: 34042038.

[19] N. Thammawongsa, F. D. Zainol, S. Mitatha, J. Ali and P. P. Yupapin, "Nanorobot Controlled by Optical Tweezer Spin for Microsurgical Use," in IEEE Transactions on Nanotechnology, vol. 12, no. 1, pp. 29-34, Jan. 2013, doi: 10.1109/TNANO.2012.2225638.

[20] Jizhuang Wang, Ze Xiong, Jing Zheng, Xiaojun Zhan, and Jinyao Tang Accounts of Chemical Research 2018 51 (9), 1957-1965 DOI: 10.1021/acs.accounts.8b00254

[21] Wu R, Zhu Y, Cai X, Wu S, Xu L, Yu T. Recent Process in Microrobots: From Propulsion to Swarming for Biomedical Applications. Micromachines (Basel). 2022 Sep 5;13(9):1473. doi: 10.3390/mi13091473. PMID: 36144096; PMCID: PMC9503943.

[22] S. Yan et al., "Ultrafast Ultrasound Imaging for Micro-Nanomotors: A Phantom Study," 2021 IEEE International Ultrasonics Symposium (IUS), Xi'an, China, 2021, pp. 1-4, doi: 10.1109/IUS52206.2021.9593877.

[23] Loscrí V, Vegni AM. An Acoustic Communication Technique of Nanorobot Swarms for Nanomedicine Applications. IEEE Trans Nanobioscience. 2015 Sep;14(6):598-607. doi: 10.1109/TNB.2015.2423373. Epub 2015 Apr 17. PMID: 25898028.

[24] Ilya L. Sokolov, Vladimir R. Cherkasov, Andrey A. Tregubov, Sveatoslav R. Buiucli, Maxim P. Nikitin,
Smart materials on the way to theranostic nanorobots: Molecular machines and nanomotors, advanced biosensors, and intelligent vehicles for drug delivery, Biochimica et Biophysica Acta (BBA) - General Subjects,Volume 1861, Issue 6,2017,Pages 1530-1544,ISSN 0304-4165, https://doi.org/10.1016/j.bbagen.2017.01.027.
(https://www.sciencedirect.com/science/article/pii/S0304416517300351)

[25] Su H, Li S, Yang GZ, Qian K. Janus Micro/Nanorobots in Biomedical Applications. Adv Healthc Mater. 2023 Jun;12(16):e2202391. doi: 10.1002/adhm.202202391. Epub 2022 Nov 29. PMID: 36377485.

[26] Li J, Dekanovsky L, Khezri B, Wu B, Zhou H, Sofer Z. Biohybrid Micro- and Nanorobots for Intelligent Drug Delivery. Cyborg Bionic Syst. 2022 Feb 10;2022:9824057. doi: 10.34133/2022/9824057. PMID: 36285309; PMCID: PMC9494704.
R. Majumdar, J. S. Rathore and N. N. Sharma, "Simulation of swimming Nanorobots in biological fluids," 2009 4th International Conference on Autonomous Robots and Agents, Wellington, New Zealand, 2009, pp. 79-82, doi: 10.1109/ICARA.2000.4803912.

[27] S. Subramanian, J. S. Rathore and N. N. Sharma, "Design and analysis of helical flagella propelled nanorobots," 2009 4th IEEE International Conference on Nano/Micro Engineered and Molecular Systems, Shenzhen, China, 2009, pp. 950-953, doi: 10.1109/NEMS.2009.5068731.

[28] Kostarelos K. Nanorobots for medicine: how close are we? Nanomedicine (Lond). 2010 Apr;5(3):341-2. doi: 10.2217/nnm.10.19. PMID: 20394527.

[29] Yang L, Zhao Y, Xu X, Xu K, Zhang M, Huang K, Kang H, Lin HC, Yang Y, Han D. An Intelligent DNA Nanorobot for Autonomous Anticoagulation. Angew Chem Int Ed Engl. 2020 Sep 28;59(40):17697-17704. doi: 10.1002/anie.202007962. Epub 2020 Aug 11. PMID: 32573062.

[30] Wang D, Li S, Zhao Z, Zhang X, Tan W. Engineering a Second-Order DNA Logic-Gated Nanorobot to Sense and Release on Live Cell Membranes for Multiplexed Diagnosis and Synergistic Therapy. Angew Chem Int Ed Engl. 2021 Jul 12;60(29):15816-15820. doi: 10.1002/anie.202103993. Epub 2021 Jun 11. PMID: 33908144.'

[31] Hu Y. Self-Assembly of DNA Molecules: Towards DNA Nanorobots for Biomedical Applications. Cyborg Bionic Syst. 2021 Oct 19;2021:9807520. doi: 10.34133/2021/9807520. PMID: 36285141; PMCID: PMC9494698.

[32] T. Hayakawa, S. Fukada and F. Arai, "Fabrication of an On-Chip Nanorobot Integrating Functional Nanomaterials for Single-Cell Punctures," in IEEE Transactions on Robotics, vol. 30, no. 1, pp. 59-67, Feb. 2014, doi: 10.1109/TRO.2013.2284402.

[33] A. Cavalcanti, "Assembly automation with evolutionary nanorobots and sensor-based control applied to nanomedicine," in IEEE Transactions on Nanotechnology, vol. 2, no. 2, pp. 82-87, June 2003, doi: 10.1109/TNANO.2003.812590.

[34] Zhang, Y.; Zhang, Y.; Han, Y.; Gong, X. Micro/Nanorobots for Medical Diagnosis and Disease Treatment. Micromachines 2022, 13, 648. https://doi.org/10.3390/mi13050648

[35] R. Hariharan and J. Manohar, "Nanorobotics as medicament: (Perfect solution for cancer)," INTERACT-2010, Chennai, India, 2010, pp. 4-7, doi: 10.1109/INTERACT.2010.5706153.

[36] Saadeh Y, Vyas D. Nanorobotic Applications in Medicine: Current Proposals and Designs. Am J Robot Surg. 2014 Jun;1(1):4-11. doi: 10.1166/ajrs.2014.1010. PMID: 26361635; PMCID: PMC4562685.

[37] Jadczyk T, Caluori G, Wojakowski W, Starek Z. Nanotechnology and stem cells in vascular biology. Vasc Biol. 2019 Sep 24;1(1):H103-H109. doi: 10.1530/VB-19-0021. PMID: 32923961; PMCID: PMC7439937.

[38] Q. Wang and L. Zhang, "Ultrasound Imaging and Tracking of Micro/Nanorobots: From Individual to Collectives," in IEEE Open Journal of Nanotechnology, vol. 1, pp. 6-17, 2020, doi: 10.1109/OJNANO.2020.2981824.

[39] Wang M, Li X, He F, Li J, Wang HH, Nie Z. Advances in Designer DNA Nanorobots Enabling Programmable Functions. Chembiochem. 2022 Sep 16;23(18):e202200119. doi: 10.1002/cbic.202200119. Epub 2022 May 16. PMID: 35491242.

[40] Genchi GG, Marino A, Grillone A, Pezzini I, Ciofani G. Remote Control of Cellular Functions: The Role of Smart Nanomaterials in the Medicine of the Future. Adv Healthc Mater. 2017 May;6(9). doi: 10.1002/adhm.201700002. Epub 2017 Mar 24. PMID: 28338285.

[41] A. Denasi and S. Misra, "Independent and Leader–Follower Control for Two Magnetic Micro-Agents," in IEEE Robotics and Automation Letters, vol. 3, no. 1, pp. 218-225, Jan. 2018, doi: 10.1109/LRA.2017.2737484.

[42] Torlakcik H, Sarica C, Bayer P, Yamamoto K, Iorio-Morin C, Hodaie M, Kalia SK, Neimat JS, Hernesniemi J, Bhatia A, Nelson BJ, Pané S, Lozano AM, Zemmar A. Magnetically Guided Catheters, Micro- and Nanorobots for Spinal Cord Stimulation. Front Neurorobot. 2021 Oct 20;15:749024. doi: 10.3389/fnbot.2021.749024. PMID: 34744678; PMCID: PMC8565609.

[43] J. Al-Sawwa and S. A. Ludwig, "Centroid-Based Particle Swarm Optimization Variant for Data Classification," 2018 IEEE Symposium Series on Computational Intelligence (SSCI), Bangalore, India, 2018, pp. 672-679, doi: 10.1109/SSCI.2018.8628926.

[44] al-Rifaie MM, Aber A, Hemanth DJ. Deploying swarm intelligence in medical imaging identifying metastasis, micro-calcifications and brain image segmentation. IET Syst Biol. 2015 Dec;9(6):234-44. doi: 10.1049/iet-syb.2015.0036. PMID: 26577158; PMCID: PMC8687301.

[45] P. Joshi, J. Leclerc, D. Bao and A. T. Becker, "Motion-planning Using RRTs for a Swarm of Robots Controlled by Global Inputs," 2019 IEEE 15th International Conference on Automation Science and Engineering (CASE), Vancouver, BC, Canada, 2019, pp. 1163-1168, doi: 10.1109/COASE.2019.8842916.

[46] L. Yang, J. Yu, S. Yang, B. Wang, B. J. Nelson and L. Zhang, "A Survey on Swarm Microrobotics," in IEEE Transactions on Robotics, vol. 38, no. 3, pp. 1531-1551, June 2022, doi: 10.1109/TRO.2021.3111788.

[47] Cavalcanti A, Shirinzadeh B, Freitas RA Jr, Kretly LC. Medical nanorobot architecture based on nanobioelectronics. Recent Pat Nanotechnol. 2007;1(1):1-10. doi: 10.2174/187221007779814745. PMID: 19076015.

[48] Jain KK. Nanomedicine: application of nanobiotechnology in medical practice. Med Princ Pract. 2008;17(2):89-101. doi: 10.1159/000112961. Epub 2008 Feb 19. PMID: 18287791.

[49] De Stefani A, Bruno G, Preo G, Gracco A. Application of Nanotechnology in Orthodontic Materials: A State-of-the-Art Review. Dent J (Basel). 2020 Nov 9;8(4):126. doi: 10.3390/dj8040126. PMID: 33182424; PMCID: PMC7712537.

[50] Jandt KD, Watts DC. Nanotechnology in dentistry: Present and future perspectives on dental nanomaterials. Dent Mater. 2020 Nov;36(11):1365-1378. doi: 10.1016/j.dental.2020.08.006. Epub 2020 Sep 25. PMID: 32981749; PMCID: PMC7516471.

[51] Shashirekha G, Jena A, Mohapatra S. Nanotechnology in Dentistry: Clinical Applications, Benefits, and Hazards. Compend Contin Educ Dent. 2017 May;38(5):e1-e4. PMID: 28459243.

[52] R. M. Merina, "Use of nanorobots in heart transplantation," INTERACT-2010, Chennai, India, 2010, pp. 265-268, doi: 10.1109/INTERACT.2010.5706155.